\documentclass[10pt,twocolumn,letterpaper]{article}

\usepackage{cvpr}
\usepackage{times}
\usepackage{epsfig}
\usepackage{graphicx}
\usepackage{amsmath}
\usepackage{amssymb}
\usepackage{subcaption}
\usepackage{media9}
\usepackage{balance}
\usepackage[font=small,skip=0pt]{caption}
\usepackage[utf8]{inputenc}
\usepackage[T1]{fontenc}
\usepackage{enumitem}
\usepackage[pagebackref=true,breaklinks=true,letterpaper=true,colorlinks,bookmarks=false]{hyperref}

\cvprfinalcopy 

\def\cvprPaperID{498} 

\ifcvprfinal\fi
\begin{document}

\title{G$^{3}$AN: Disentangling Appearance and Motion for Video Generation}

\author{Yaohui Wang\textsuperscript{1,2} \hskip 1em 
Piotr Bilinski\textsuperscript{3} \hskip 1em 
Francois Bremond\textsuperscript{1,2} \hskip 1em 
Antitza Dantcheva\textsuperscript{1,2}\\
\textsuperscript{1}Inria \hskip 1em
\textsuperscript{2}Université Côte d'Azur \hskip 1em
\textsuperscript{3}University of Warsaw\\
{\tt\small \{yaohui.wang, francois.bremond, antitza.dantcheva\}@inria.fr} \hskip 1em
{\tt\small bilinski@mimuw.edu.pl}
}

\maketitle
\thispagestyle{empty}

\begin{abstract}
Creating realistic human videos entails the challenge of being able to simultaneously generate both appearance, as well as motion.
To tackle this challenge, we introduce \textsc{G$^{3}$AN}, a novel spatio-temporal generative model, which seeks to capture the distribution of high dimensional video data and to model appearance and motion in disentangled manner.
The latter is achieved by decomposing appearance and motion in a three-stream Generator, where the main stream aims to model spatio-temporal consistency, whereas the two auxiliary streams augment the main stream with multi-scale appearance and motion features, respectively.
An extensive quantitative and qualitative analysis shows that our model systematically and significantly outperforms state-of-the-art methods on the facial expression datasets MUG and UvA-NEMO, as well as the Weizmann and UCF101 datasets on human action.
Additional analysis on the learned latent representations confirms the successful decomposition of appearance and motion. Source code and pre-trained models are publicly available \footnote{\url{https://wyhsirius.github.io/G3AN/}}.
\end{abstract}

\section{Introduction}
Generative Adversarial Networks (GANs)~\cite{goodfellow2014generative} have witnessed increasing attention due to their ability to model complex data distributions, which allows them to \textit{generate} realistic \textit{images} \cite{brock2018large,karras2017progressive,karras2019style,ledig2017photo, ma2018disentangled,miyato2018spectral,xu2018attngan,zhao2019image}, as well as to translate images \cite{albahar2019guided,isola2017image,romero2019smit,siddiquee2019learning}. While realistic \textit{video generation} is the natural sequel, it is substantially more challenging \wrt complexity and computation, associated to the simultaneous modeling of appearance, as well as motion.

Specifically, in inferring and modeling the distribution of human videos, generative models face three main challenges: (a) generating uncertain motion, (b) retaining of human appearance throughout the generated video, as well as (c) modeling spatio-temporal consistency. Such challenges have been alleviated by conditioning the generation on potent priors such as input images~\cite{yang2018pose}, human keypoints~\cite{chan2019everybody} and optical flows~\cite{li2018flow}. This relates to learning to sample from conditional distributions, assuming access to the marginal distributions instead of learning joint distributions~\cite{pu2018jointgan}.

Deviating from such approaches, in this work we focus on the highly intricate problem of video generation without prior knowledge \wrt either appearance or motion.
Specifically, based on noise variables, we generate an appearance, \eg human face and body, which we concurrently animate, by a facial expression or human action.

G$^{3}$AN, our new generative model, is streamlined to learn a \textit{disentangled representation} of the video generative factors \textit{appearance} and \textit{motion}, allowing for manipulation of both.
A disentangled representation has been defined as one, where single latent units are sensitive to changes in single generative factors, while being relatively invariant to changes in other factors \cite{bengio2013representation}. In this context, our G$^{3}$AN is endowed with a three-stream Generator-architecture, where the main stream encodes spatio-temporal video representation, augmented by two auxiliary streams, representing the independent generative factors \textit{appearance} and \textit{motion}. A self-attention mechanism targeted towards high level feature maps ensures satisfactory video quality.

G$^{3}$AN is hence able to generate realistic videos (tackling challenges (a) and (c)) by following a training distribution and without additional input, as well as is able to manipulate the appearance and motion disjointly, while placing emphasis on preserving appearance (challenge (b)).

In summary, our \textbf{main technical contributions} include the following.
\setlist{nolistsep}
\begin{itemize}[noitemsep]
\item A novel generative model, G$^{3}$AN, which seeks to learn disentangled representations of the generative factors \textit{appearance} and \textit{motion} from human video data. The representations allow for individual \textit{manipulation} of both factors. 
\item A novel three-stream generator, which takes into account the learning of individual appearance features (\textit{spatial stream}), motion features (\textit{temporal stream}) and smoothing generated videos (\textit{main stream}) at the same time.
\item A novel \textit{factorized spatio-temporal self-attention (F-SA)}, which is considered as the first self-attention module applied to video generation, in order to model global spatio-temporal representations and improve the quality of generated videos.
\item Extensive qualitative and quantitative evaluation, which demonstrates that G$^{3}$AN systematically and significantly outperforms state-of-the-art baselines on a set of datasets. 
\end{itemize}

\section{Related Work}\label{rel_work}
Despite the impressive progress of image generation, the extension to \textit{video} generation is surprisingly challenging. While videos constitute sequences of temporally coherent images, video generation encompasses a majority of challenges that have to do with generation of plausible and realistic appearance, coherent and realistic motion, as well as spatio-temporal consistency. A further challenge, namely the generation of uncertain local or global motion, associated to future uncertainty, allows for multiple correct, equally probable next frames \cite{walker2016uncertain}. 
Finding  suitable  representation learning methods, which are able to address these challenges is critical. Existing methods include approaches based on Variational Autoencoders (VAEs) \cite{kingma2013auto}, auto-regressive models, as well as most prominently Generative Adversarial Networks (GANs)~\cite{goodfellow2014generative}. 

While video generation tasks aim at generating realistic temporal dynamics, such tasks vary with the \textit{level of conditioning}. We have video generation based on additional priors related to motion or appearance, as well as contrarily, video generation following merely the training distribution. 
We note that the latter is more challenging from a modeling perspective, due to lack of additional input concerning \eg structure of the generated video. Therefore the majority of approaches to date include a conditioning of some kind. 

\textbf{Video generation with additional input.} Due to challenges in modeling of high dimensional video data, additional information such as semantic maps~\cite{pan2019video, wang2018vid2vid}, human keypoints \cite{jang2018video, yang2018pose, walker2017pose, chan2019everybody}, 3D face mesh \cite{Zhao_2018_ECCV} and optical flow \cite{li2018flow} can be instrumental as guidance for appearance and motion generation. This additional information is either pre-computed throughout the generated video \cite{jang2018video, Zhao_2018_ECCV, chan2019everybody} or predicted based on an initial input image \cite{yang2018pose}. The additional information guides conditional image translation, which though results in lack of modeling of spatio-temporal correlations.

\textbf{Video generation from noise.} Directly generating videos from noise requires the capturing and modeling of a dataset distribution. Existing works tend to reduce related complexity by decomposing either the output~\cite{vondrick2016generating} or latent representation~\cite{saito2017temporal, tulyakov2017mocogan}. VGAN~\cite{vondrick2016generating} was equipped with a two-stream spatio-temporal Generator, generating foreground and background separately. TGAN~\cite{saito2017temporal} decomposed the latent representation of each frame into a \textit{slow part} and a \textit{fast part}. Due to jointly modeling appearance and motion, generated results from VGAN and TGAN might comprise spatially unrealistic artefacts, see Figure~\ref{fig:comparison}. 
The closest work to ours is MoCoGAN~\cite{tulyakov2017mocogan}, which decomposed the latent representation of \textit{each frame} into motion and content, aiming at controlling both factors. However, there are two crucial differences between MoCoGAN and G$^{3}$AN. Firstly, instead of only sampling two noise vectors for each video, MoCoGAN sampled a sequence of noise vectors as motion and a fixed noise as content. However, involving random noise for each frame to represent motion increases the learning difficulty, since the model has to map these noise vectors to a consecutive human movement in the generated videos. As a result, MoCoGAN gradually ignores the input noise and tends to produce a similar motion, as we illustrate in Figure~\ref{fig:mug_us_mocogan}. 
Secondly, MoCoGAN incorporated a simple image Generator aiming at generating each frame sequentially, after which content and motion features were jointly generated. This leads to \textit{incomplete disentanglement} of motion and content. Deviating from that, we design a novel Generator architecture, able to entirely decompose appearance and motion in both, latent and feature spaces. We show that such design generates realistic videos of good quality and ensures factor disentanglement.

\textbf{Disentangled representation learning.}
Learning disentangled representations of data has been beneficial in a large variety of tasks and domains~\cite{bengio2013representation}. Disentangling a number of factors in \textit{still images} has been widely explored in recent works~\cite{chen2016infogan, ma2018disentangled, Singh_2019_CVPR, Lee_2018_ECCV}. In the context of \textit{video generation}, an early approach for motion and appearance decomposition was incorporated in MoCoGAN. However, experiments, which we present later (see Figure~\ref{fig: uva_uncond}), suggest that the results are not satisfactory.

\begin{figure*}[htbp]
\centering
\includegraphics[width=1\textwidth]{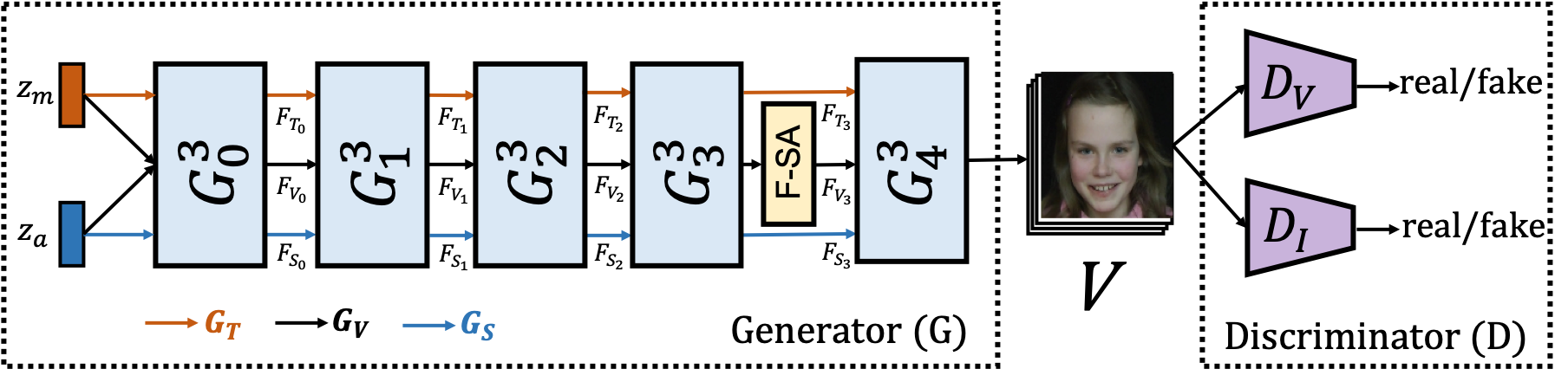}  
\caption{\textbf{Overview of our G$^{3}$AN architecture.} G$^{3}$AN consists of a three-stream Generator and a two-stream Discriminator. The Generator contains five stacked G$^3$ modules, a factorized self-attention (F-SA) module, and takes as input two random noise vectors, $z_a$ and $z_m$, aiming at representing appearance and motion, respectively. Details of architecture can be found in Supplementary Material (SM).}
\label{fig:architecture}
\end{figure*}

\section{Approach}\label{proposed_method}
In this work, we propose G$^{3}$AN, a novel GAN architecture, aiming at generating videos in a disentangled manner from two noise vectors, $z_{a}\in Z_{A}$ and $z_{m} \in Z_{M}$, which represent appearance and motion, respectively. G$^{3}$AN consists of a three-stream Generator $G$, as well as a two-stream Discriminator $D$, as illustrated in Figure~\ref{fig:architecture}. While $G$ aims at generating videos with the ability to modulate appearance and motion disjointly, $D$ accounts for distinguishing generated samples from real data, in both, videos and frames, respectively.

\subsection{Generator}
\begin{figure}[htb]
\centering
\includegraphics[width=0.47\textwidth]{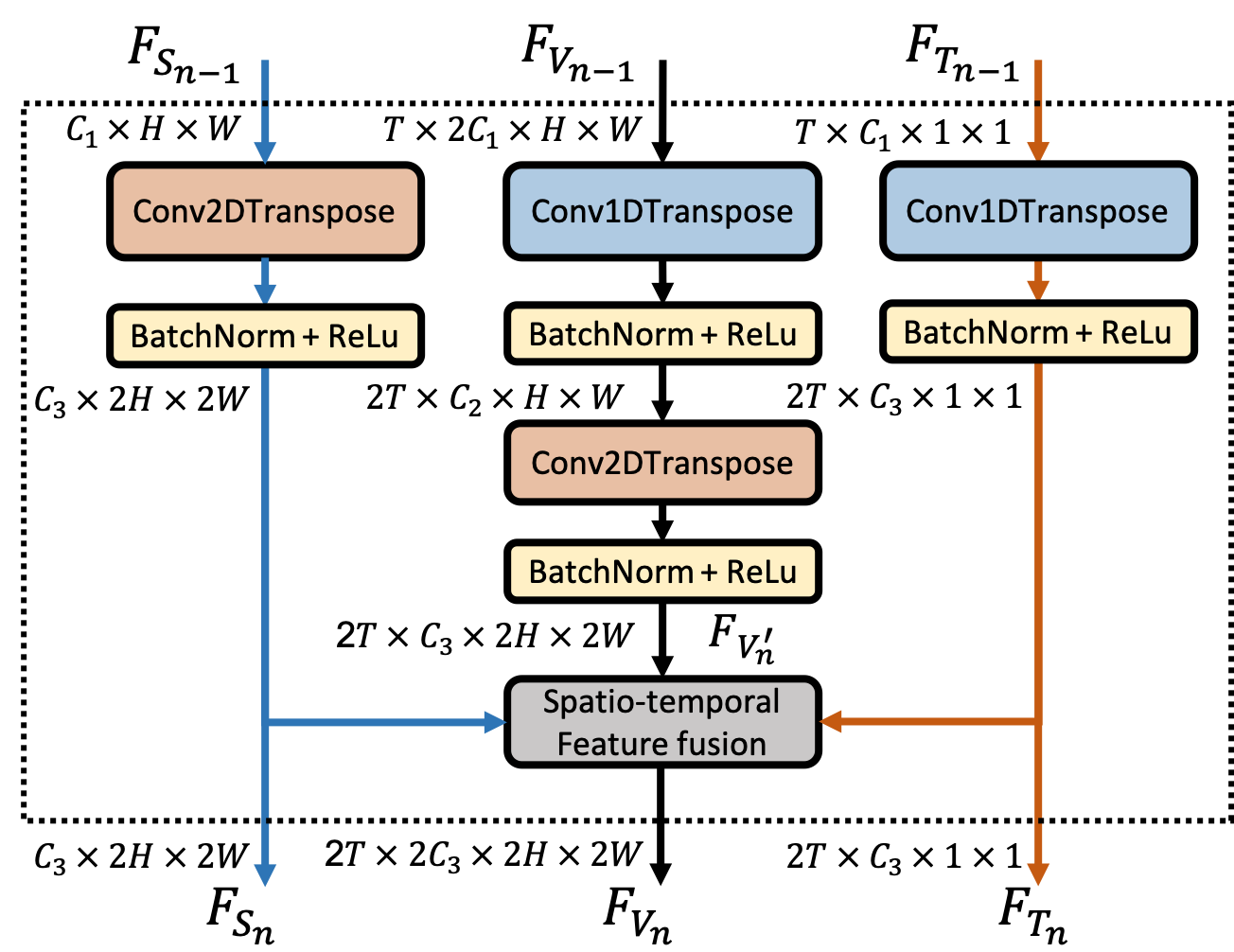} 
\caption{\textbf{G$^3$ module architecture.}}
\label{fig:g3_module}
\end{figure}

\textbf{Hierarchical Generator with G$^{3}$-modules}.
We design $G$ in a hierarchical structure of G$^3$ modules. Specifically, we have $N$ levels of hierarchy, denoted as G$_{n=0...N-1}^3$.
The first G$^3$ module, G$_0^3$ accepts as input the two noise vectors $z_a$ and $z_m$.
The remaining modules G$_{n=1...N-1}^3$, inherit the three feature maps $F_{S_{n-1}}$, $F_{V_{n-1}}$ and $F_{T_{n-1}}$ as their inputs from each previous G$^3_{n-1}$ module, see Figures~\ref{fig:architecture} and~\ref{fig:g3_module}.

Each G$_n^3$ module consists of three parallel streams: a spatial stream $G_{S_n}$, a temporal stream $G_{T_n}$, as well as a video stream $G_{V_n}$ (Figures~\ref{fig:architecture} and~\ref{fig:g3_module}). They are designed to generate three different types of features. The spatial stream $G_{S_n}$, denoted by a blue line in Figures~\ref{fig:architecture} and~\ref{fig:g3_module}, takes as input $z_a$ for $n=0$ and $F_{S_{n-1}}$ for $n>1$, and generates 2D appearance features $F_{S_n}$ by upsampling input features with a transposed 2D convolutional layer. These features evolve in spatial dimension and are shared at all time instances. The temporal stream $G_{T_n}$, denoted by an orange line, accepts as input $z_m$ for $n=0$ and $F_{T_{n-1}}$ for $n>1$, and seeks to generate 1D motion features $F_{T_n}$ by upsampling input features with a transposed 1D convolutional layer. These features evolve in temporal dimension and contain global information of each time step. Then, the video stream $G_{V_n}$, denoted by a black line, takes as input the concatenation of $z_a$ and $z_m$ for $n=0$ and $F_{V_{n-1}}$ for $n>1$.
It models spatio-temporal consistency and produces 3D joint embeddings $F_{V_n^{'}}$ by upsampling input features with a factorized transposed spatio-temporal convolution, see below.
Then, $F_{S_{n}}$ and $F_{T_{n}}$ are catapulted to the spatio-temporal fusion block, where they are fused with $F_{V_{n}^{'}}$, resulting in $F_{V_{n}}$. Finally, $F_{S_n}$, $F_{T_n}$ and $F_{V_n}$ serve as inputs of the next hierarchy-layer G$_{n+1}^3$.

\textbf{Factorized transposed spatio-temporal convolution} has been proposed by Wang \etal in~\cite{wang:hal-02368319}. It explicitly factorizes transposed 3D convolution into two separate and successive operations, $M$ transposed 1D temporal convolution followed by a 2D separate spatial convolution, which is referred to as transposed (1+2)D convolution. Such decomposition brings an additional nonlinear activation between these two operations and facilitates optimization. Crucially, factorizing transposed 3D convolution yields significant gains in video quality, see Section \ref{lab:experiments}.

\begin{figure}[htb]
\centering
\includegraphics[width=0.47\textwidth]{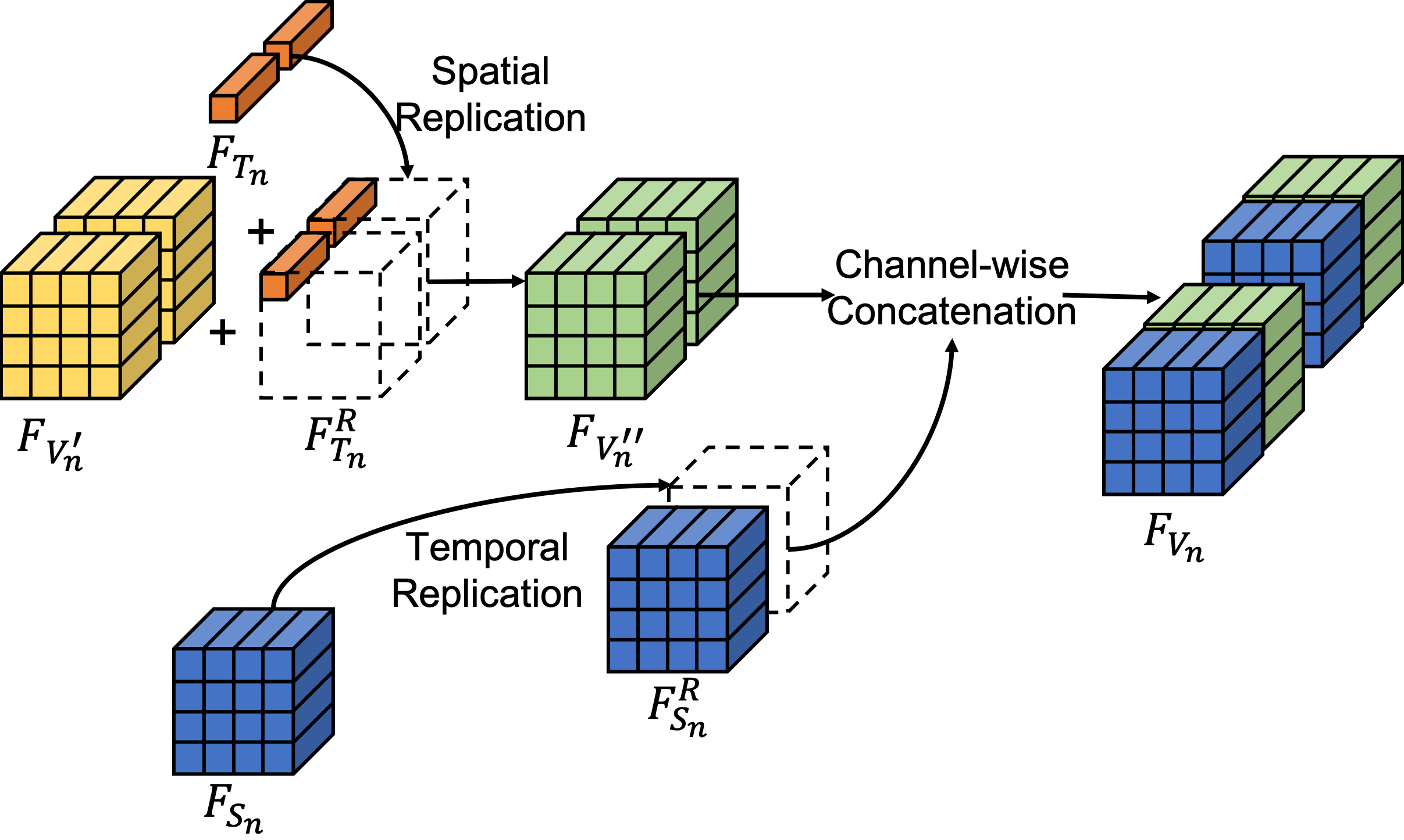}
\caption{\textbf{Spatio-temporal fusion.}} 
\label{fig:fusion}
\end{figure}

\textbf{Spatio-temporal fusion} is the key-element 
to learn well disentangled features, 
the inputs of which 
are output feature maps $F_{S_n}$, $F_{T_n}$ and $F_{V^{'}_{n}}$ from the convolutional layers in each G$^{3}_{n}$ module. The fusion contains three steps (see Figure~\ref{fig:fusion}). Firstly, spatial and temporal replications are applied on $F_{T_n}$ and $F_{S_n}$ respectively,
in order to obtain two new feature maps $F^R_{T_n}$ and $F^R_{S_n}$.
Both new feature maps have the same spatio-temporal size as $F_{V^{'}_n}$. Next, $F^R_{T_n}$ and $F_{V^{'}_n}$ are combined through a position-wise addition, creating a new spatio-temporal embedding $F_{V^{''}_n}$. Finally, $F^R_{S_n}$ is channel-wise concatenated with $F_{V^{''}_n}$, obtaining the final fused feature map $F_{V_n}$. The feature maps $F_{S_n}$, $F_{T_n}$ and $F_{V_n}$ represent inputs for the following G$^3_{n+1}$ module.

\begin{figure}[htb]
\centering
\includegraphics[width=7.5cm]{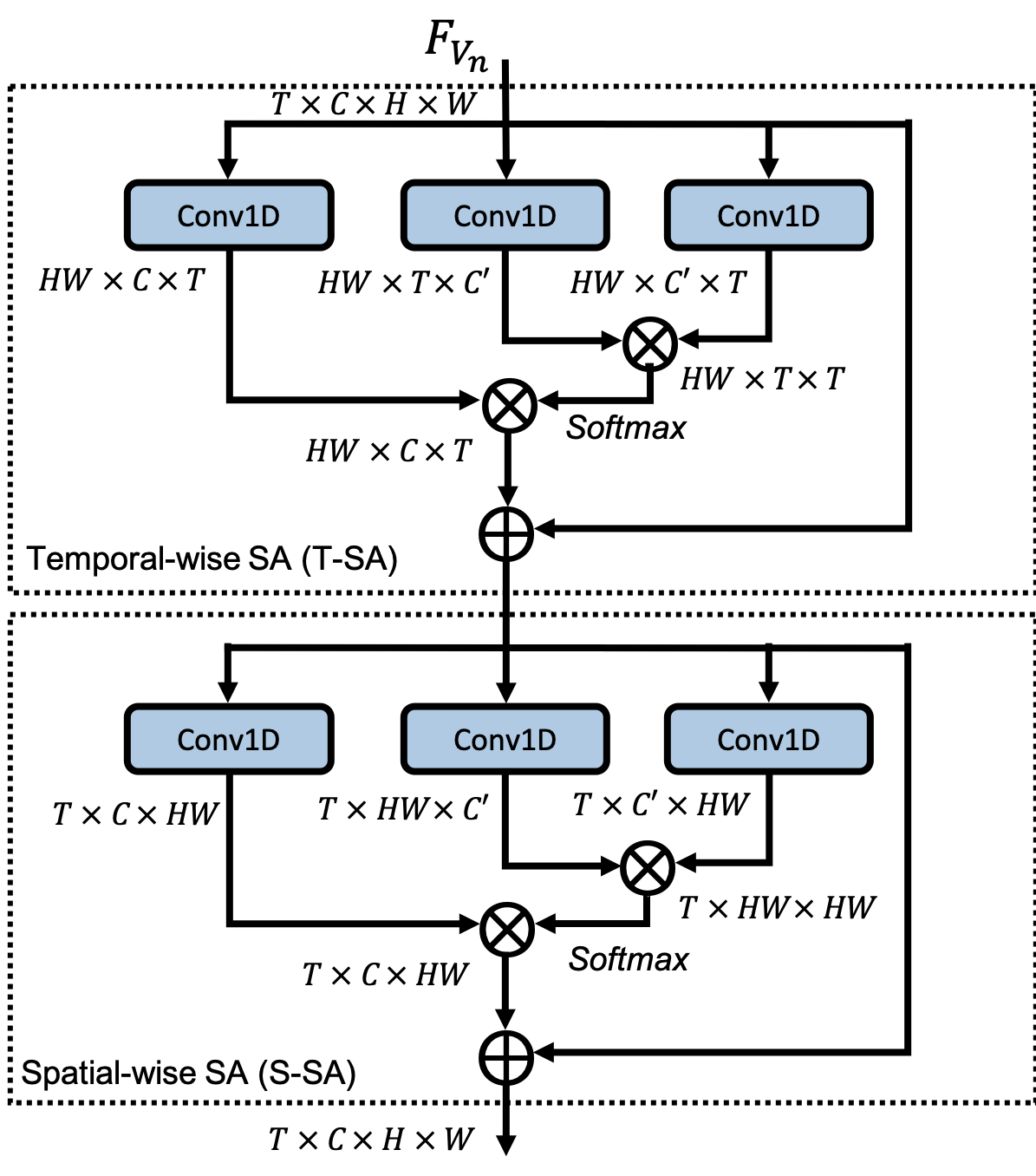}  
\caption{\textbf{Factorized spatio-temporal Self-Attention (F-SA).}}
\label{fig:self_attention}
\end{figure}

\textbf{Factorized spatio-temporal Self-Attention (F-SA)}.
Despite Self-Attention (SA) has been successfully applied in image generation within SAGAN~\cite{zhang2018self}, however it has not been explored in the context of spatio-temporal video generation yet. Here, we incorporate a spatio-temporal SA module, enabling $G$ to utilize cues from all spatio-temporal feature positions and model relationships between widely separated regions. However, computing correlation between each position with all the others in a 3D spatio-temporal feature map is very computationally expensive, particularly if it is applied on higher feature maps in $G$.
Therefore, we propose a novel \textit{factorized spatio-temporal self-attention}, namely F-SA, as shown in Figure~\ref{fig:self_attention}. F-SA consists of a Temporal-wise SA (T-SA), followed by a Spatial-wise SA (S-SA). Such factorization reduces the computational complexity, allowing for application of the F-SA on larger feature maps.

In our G$^{3}$AN, we apply F-SA on the output of the G$^3_{3}$ in the $G_V$ stream, which achieve best video quality. We report related evaluation results of applying F-SA at various hierarchy-layers of the G$^{3}$AN in Section~\ref{lab:experiments}.

\subsection{Discriminator}
Towards improving both video and frame quality,  similar to MoCoGAN, we use a two-stream \textit{Discriminator} architecture, containing a video stream $D_{V}$ and an image stream $D_{I}$. During training, $D_{V}$ accepts a full video as input, whereas $D_{I}$ takes randomly sampled frames from videos.

\subsection{Training}
Given our two-stream Discriminator architecture, G$^{3}$AN simultaneously optimizes $D_V$ and $D_I$. Both losses use the GAN loss function proposed in DCGAN~\cite{radford2015unsupervised}. The objective functions of G$^{3}$AN can be expressed as

\begin{alignat}{2}
G^* = arg\min_G \max_{D_I, D_V} L(G, D_I, D_V), ~~~~~~~~\\
L(G, D_I, D_V) = L_{I}(G,D_I) + L_{V}(G,D_V), ~~~~~~~~
\end{alignat}
\noindent where $L_{I}$ denotes the loss function related to $D_{I}$, $L_{V}$ represents the loss function related to $D_{V}$. 
\begin{alignat}{2}
L_{I} &&= & ~ \mathbb{E}_{x^{'} \sim p_{data}}[log(D_{I}(x^{'}))] \nonumber \\
&&+ & ~ \mathbb{E}_{z_a \sim p_{z_a}, z_m \sim p_{z_m}}[log(1-D_{I}(G(z_{a},z_{m})^{'}))],~~\\
L_{V} &&= & ~ \mathbb{E}_{x \sim p_{data}}[log(D_{V}(x))] \nonumber \\
&&+ & ~ \mathbb{E}_{z_a \sim p_{z_a}, z_m \sim p_{z_m}}[log(1-D_{V}(G(z_{a},z_{m})))], 
\end{alignat}

\noindent $G$ attempts to generate videos from $z_a$ and $z_m$, while $D_I$ and $D_V$ aim to distinguish between generated samples and real samples. $(\cdot)^{'}$ characterizes that $T$ frames are being sampled from real and generated videos. 




\section{Experiments}\label{lab:experiments}
\subsection{Implementation details}
The entire network is implemented using PyTorch.
We employ ADAM optimizer~\cite{kingma2014adam} with $\beta_{1}{=}0.5$ and $\beta_{2}{=}0.999$.
Learning rate is set to $2e^{-4}$ for both $G$ and $D$.
Dimensions of latent representations constitute $128$ for $z_a$ and $10$ for $z_m$. We set $N=5$ in order to generate videos of 16 frames with spatial scale $64\times 64$. We randomly sample $T=1$ frame from each video as input of $D_I$. More implementation details can be found in SM.

\subsection{Datasets}
We evaluate our method on following four datasets.

\textbf{Facial expression datasets.} The \textbf{MUG} Facial Expression dataset~\cite{aifanti2010mug} contains 1254 videos of 86 subjects, performing 6 facial expressions, namely \textit{happy}, \textit{sad}, \textit{surprise}, \textit{anger}, \textit{disgust} and \textit{fear}. The \textbf{UvA-NEMO} Smile dataset~\cite{dibekliouglu2012you} comprises 1240 video sequences of 400 smiling individuals, with 1 or 2 videos per subject. We crop faces in each frame based on detected landmarks using~\cite{bulat2017far}$\footnote{\url{https://github.com/1adrianb/face-alignment}}$.


\textbf{Action recognition datasets.} The \textbf{Weizmann} Action dataset~\cite{ActionsAsSpaceTimeShapes_pami07} consists of videos of 9 subjects, performing 10 actions such as \textit{wave} and \textit{bend}. We augment it by horizontally flipping the existing videos. The \textbf{UCF101} dataset \cite{soomro2012ucf101} contains 13,320 videos of 101 human action classes. Similarly to TGAN~\cite{saito2017temporal}, we scale each frame to $85\times64$ and crop the central $64\times64$ regions.

\indent In all our experiments, we sample video frames with a random time step ranging between $1$ and $4$ for data augmentation.

\subsection{Experimental Results}
We test our method both quantitatively and qualitatively, providing results of four experiments. Specifically, firstly we
evaluate and compare videos generated from G$^{3}$AN, VGAN, TGAN and MoCoGAN, quantitatively and qualitatively on all four datasets.
Next, we test \textit{conditional} and \textit{unconditional} video generation, where we aim to demonstrate the effectiveness of the proposed decomposition method. Then, we manipulate the latent representation, providing insight into each dimension of the two representations. We proceed to add appearance vectors and study the latent representation.   Finally, we conduct an ablation study, verifying the effectiveness of our proposed architecture.

\subsubsection{Quantitative Evaluation}
We compare G$^{3}$AN with three state-of-the-art methods, namely VGAN, TGAN, as well as MoCoGAN. We report two evaluation metrics on the above four datasets.
In particular, we use the extension of two most commonly used metrics in image generation, the Inception Score (IS) \cite{NIPS2016_6125} and Fr\'echet Inception Distance (FID) \cite{NIPS2017_7240}, in video level by using a pre-trained 3D CNN~\cite{hara2018can} as feature extractor, similar to Wang~\etal~\cite{wang2018vid2vid}. 

The video \textbf{FID} grasps both visual quality and temporal consistency of generated videos. It is calculated as $\|\mu - \widetilde{\mu} \|^{2} + Tr(\Sigma + \widetilde{\Sigma} - 2\sqrt{\Sigma\widetilde{\Sigma}})$, where $\mu$ and $\Sigma$ represent the mean and covariance matrix, computed from real feature vectors, respectively, and $\widetilde{\mu}$, and $\widetilde{\Sigma}$ are computed from generated data. Lower FID scores indicate a superior quality of generated videos.

The video \textbf{IS} captures the quality and diversity of generated videos. It is calculated as $\exp(\mathbb{E}_{x\sim p_{g}} D_{KL} (p(y|x)\|p(y)))$, where $p(y|x)$ and $p(y)$ denote conditional class distribution and marginal class distribution, respectively. A higher IS indicates better model performance.

We report FID on MUG, UVA-Nemo and Weizmann datasets, and both FID and IS on UCF101. Since IS can only be reported, when GAN and feature extractor are trained on the same dataset, we do not report it on other datasets.


\begin{figure*}
\begin{subfigure}{0.247\textwidth}
\centering
\includegraphics[width=0.31\textwidth]{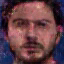}
\includegraphics[width=0.31\textwidth]{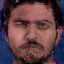}
\includegraphics[width=0.31\textwidth]{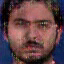}
\includegraphics[width=0.31\textwidth]{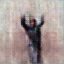}
\includegraphics[width=0.31\textwidth]{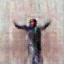}
\includegraphics[width=0.31\textwidth]{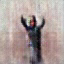}
\includegraphics[width=0.31\textwidth]{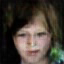}
\includegraphics[width=0.31\textwidth]{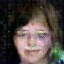}
\includegraphics[width=0.31\textwidth]{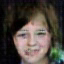}
\includegraphics[width=0.31\textwidth]{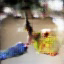}
\includegraphics[width=0.31\textwidth]{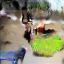}
\includegraphics[width=0.31\textwidth]{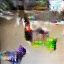}
\caption{\textbf{VGAN}}
\label{subfig:vgan}
\end{subfigure}
\begin{subfigure}{0.247\textwidth}
\centering
\includegraphics[width=0.31\textwidth]{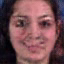}
\includegraphics[width=0.31\textwidth]{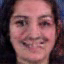}
\includegraphics[width=0.31\textwidth]{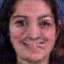}
\includegraphics[width=0.31\textwidth]{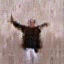}
\includegraphics[width=0.31\textwidth]{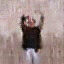}
\includegraphics[width=0.31\textwidth]{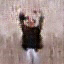}
\includegraphics[width=0.31\textwidth]{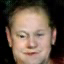}
\includegraphics[width=0.31\textwidth]{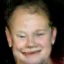}
\includegraphics[width=0.31\textwidth]{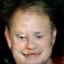}
\includegraphics[width=0.31\textwidth]{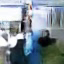}
\includegraphics[width=0.31\textwidth]{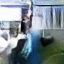}
\includegraphics[width=0.31\textwidth]{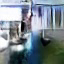}
\caption{\textbf{TGAN}}
\label{subfig:tgan}
\end{subfigure}
\begin{subfigure}{0.247\textwidth}
\centering
\includegraphics[width=0.31\textwidth]{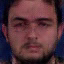}
\includegraphics[width=0.31\textwidth]{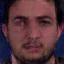}
\includegraphics[width=0.31\textwidth]{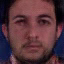}
\includegraphics[width=0.31\textwidth]{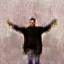}
\includegraphics[width=0.31\textwidth]{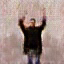}
\includegraphics[width=0.31\textwidth]{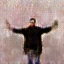}
\includegraphics[width=0.31\textwidth]{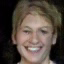}
\includegraphics[width=0.31\textwidth]{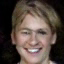}
\includegraphics[width=0.31\textwidth]{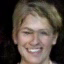}
\includegraphics[width=0.31\textwidth]{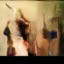}
\includegraphics[width=0.31\textwidth]{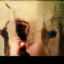}
\includegraphics[width=0.31\textwidth]{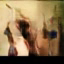}
\caption{\textbf{MoCoGAN}}
\label{subfig:mocogan}
\end{subfigure}
\begin{subfigure}{0.247\textwidth}
\centering
\includegraphics[width=0.31\textwidth]{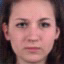}
\includegraphics[width=0.31\textwidth]{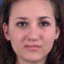}
\includegraphics[width=0.31\textwidth]{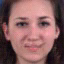}
\includegraphics[width=0.31\textwidth]{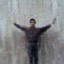}
\includegraphics[width=0.31\textwidth]{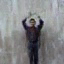}
\includegraphics[width=0.31\textwidth]{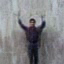}
\includegraphics[width=0.31\textwidth]{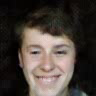}
\includegraphics[width=0.31\textwidth]{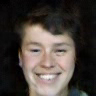}
\includegraphics[width=0.31\textwidth]{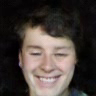}
\includegraphics[width=0.31\textwidth]{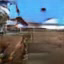}
\includegraphics[width=0.31\textwidth]{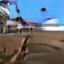}
\includegraphics[width=0.31\textwidth]{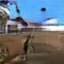}
\caption{\textbf{G$^{3}$AN}}
\label{subfig:g3an}
\end{subfigure}
\caption{\textbf{Comparison with the state-of-the-art} on MUG (top-left), Weizmann (top-right), UvA-NEMO (bottom-left) and UCF101 (bottom-right). More samples are presented in SM.}
\label{fig:comparison}
\end{figure*}

Comparison results among different methods are presented in Table \ref{tab:fid}. Our method consistently achieves the lowest FID on all datasets, suggesting that videos generated by G$^{3}$AN entail both, best temporal consistency and visual quality. At the same time, the obtained highest IS on UCF101 indicates that our method is able to provide the most diverse samples among all compared methods. Such evaluation results show that proposed decomposition method allows for controlling the generated samples, and additionally facilitates the spatio-temporal learning of generating better quality videos. Generated samples are illustrated in Figure~\ref{fig:comparison}.

\begin{table}[thb]
\centering
\setlength{\tabcolsep}{3.5pt}
\begin{tabular}{cccccc}
\hline
& {\textbf{MUG}} &
{\textbf{UvA}} &
{\textbf{Weizmann}} &
\multicolumn{2}{c}{\textbf{UCF101}} \\
& FID $\downarrow$ & FID $\downarrow$ & FID $\downarrow$ & FID $\downarrow$ & IS $\uparrow$ \\
\hline
VGAN & 160.76 & 235.01 & 158.04 & 115.06 & 2.94 \\
TGAN & 97.07 & 216.41 & 99.85 & 110.58 & 2.74 \\
MoCoGAN & 87.11 & 197.32 & 92.18 & 104.14 & 3.06 \\
G$^{3}$AN & \textbf{67.12} & \textbf{119.22} & \textbf{86.01} & \textbf{91.21} & \textbf{3.62} \\
\hline
\end{tabular}
\caption{\textbf{Comparison with the state-of-the-art} on four datasets \wrt FID and IS.}\label{tab:fid}
\end{table}

In addition, we conduct a subjective analysis, 
where we asked $27$ human raters to pairwise compare videos of pertaining to the same expression/action, displayed side by side. Raters selected one video per video-pair. We randomized the order of displayed pairs. We had an equal amount of pairs for each studied case (\eg G$^{3}$AN / Real videos). The posed question was ''Which video clip is more \textbf{realistic}$\text{?}$``. We report the mean user preference in Table \ref{tab:mup}.

We observe that human raters express a strong preference for the proposed framework G$^{3}$AN over MoCoGAN (84.26\% \vs 15.74\%), TGAN (87.31\% \vs 12.69\%) and VGAN (90.24\% \vs 9.76\%), which is consistent with the above listed quantitative results.
Further, we compare real videos from all datasets with the generated video sequences from our method.
The human raters ranked 25.71\% of videos from our G$^{3}$AN as more realistic than real videos, which we find highly encouraging.

\begin{table}[thb]
\centering
\begin{tabular}{cc}
\hline
\textbf{Methods} & \textbf{Rater preference (\%)} \\
\hline
G$^{3}$AN / MoCoGAN & \textbf{84.26} / 15.74 \\
G$^{3}$AN / TGAN  & \textbf{87.31} / 12.69 \\
G$^{3}$AN / VGAN  & \textbf{90.24} / 9.76\\
G$^{3}$AN / Real videos & 25.71 /  \textbf{74.29}\\
\hline
\end{tabular}
\caption{Mean user preference of human raters comparing videos generated by the respective algorithms, originated from all datasets.}\label{tab:mup}
\end{table}

\begin{figure}[htbp]
\begin{subfigure}{0.48\textwidth}
\centering
\includegraphics[width=0.157\textwidth]{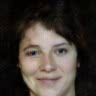}
\includegraphics[width=0.157\textwidth]{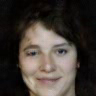}
\includegraphics[width=0.157\textwidth]{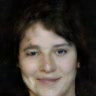}
\includegraphics[width=0.157\textwidth]{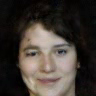}
\includegraphics[width=0.157\textwidth]{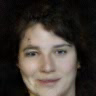}
\includegraphics[width=0.157\textwidth]{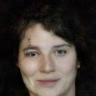}
\includegraphics[width=0.157\textwidth]{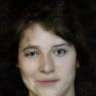}
\includegraphics[width=0.157\textwidth]{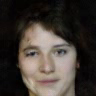}
\includegraphics[width=0.157\textwidth]{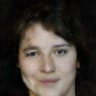}
\includegraphics[width=0.157\textwidth]{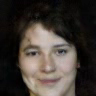}
\includegraphics[width=0.157\textwidth]{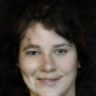}
\includegraphics[width=0.157\textwidth]{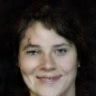}
\caption{\textbf{G$^3$AN}}
\label{subfig:uncond_ours}
\end{subfigure}
\begin{subfigure}{0.48\textwidth}
\centering
\includegraphics[width=0.157\textwidth]{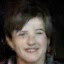}
\includegraphics[width=0.157\textwidth]{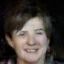}
\includegraphics[width=0.157\textwidth]{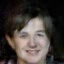}
\includegraphics[width=0.157\textwidth]{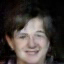}
\includegraphics[width=0.157\textwidth]{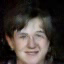}
\includegraphics[width=0.157\textwidth]{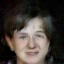}
\includegraphics[width=0.157\textwidth]{mocogan_demo1/sub1/frames_0001.png}
\includegraphics[width=0.157\textwidth]{mocogan_demo1/sub1/frames_0004.png}
\includegraphics[width=0.157\textwidth]{mocogan_demo1/sub1/frames_0007.png}
\includegraphics[width=0.157\textwidth]{mocogan_demo1/sub1/frames_0010.png}
\includegraphics[width=0.157\textwidth]{mocogan_demo1/sub1/frames_0013.png}
\includegraphics[width=0.157\textwidth]{mocogan_demo1/sub1/frames_0016.png}
\caption{\textbf{MoCoGAN}}
\label{subfig:uncond_mocogan}
\end{subfigure}
\caption{\textbf{Unconditional video generation} of G$^{3}$AN and MoCoGAN on Uva-Nemo. For each model, we fix $z_a$, while testing two $z_m$ instances (top and bottom lines). See SM for more samples.}
\label{fig: uva_uncond}
\end{figure}

\subsubsection{Qualitative Evaluation}
We conduct an \textbf{unconditional generation} experiment utilizing the Uva-NEMO dataset, where we fix $z_a$ and proceed to randomly vary motion, $z_m$. Associated generated samples from G$^{3}$AN and MoCoGAN are shown in Figure~\ref{fig: uva_uncond}.  
While we observe the varying motion in the video sequences generated by G$^{3}$AN, the appearance remains coherent. Hence, our model is able to successfully preserve facial appearance, while \textit{altering} the motion. Therefore, this suggests that our three-stream design allows for manipulation of appearance and motion separately. On the contrary, video sequences generated by MoCoGAN experience constant motion, despite of altering $z_m$. 

\begin{figure}[htbp]
\begin{subfigure}{0.48\textwidth}
\centering
\includegraphics[width=0.157\textwidth]{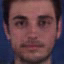}
\includegraphics[width=0.157\textwidth]{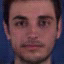}
\includegraphics[width=0.157\textwidth]{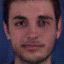}
\includegraphics[width=0.157\textwidth]{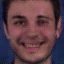}
\includegraphics[width=0.157\textwidth]{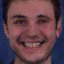}
\includegraphics[width=0.157\textwidth]{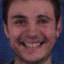}
\includegraphics[width=0.157\textwidth]{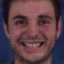}
\includegraphics[width=0.157\textwidth]{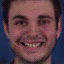}
\includegraphics[width=0.157\textwidth]{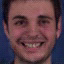}
\includegraphics[width=0.157\textwidth]{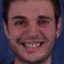}
\includegraphics[width=0.157\textwidth]{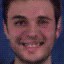}
\includegraphics[width=0.157\textwidth]{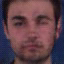}
\caption{\textbf{MUG}: \textit{Happiness}}
\label{fig:mug_cat1}
\end{subfigure}
\begin{subfigure}{0.48\textwidth}
\centering
\includegraphics[width=0.157\textwidth]{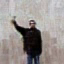}
\includegraphics[width=0.157\textwidth]{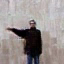}
\includegraphics[width=0.157\textwidth]{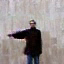}
\includegraphics[width=0.157\textwidth]{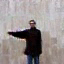}
\includegraphics[width=0.157\textwidth]{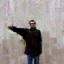}
\includegraphics[width=0.157\textwidth]{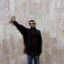}
\includegraphics[width=0.157\textwidth]{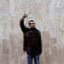}
\includegraphics[width=0.157\textwidth]{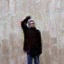}
\includegraphics[width=0.157\textwidth]{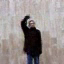}
\includegraphics[width=0.157\textwidth]{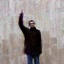}
\includegraphics[width=0.157\textwidth]{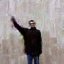}
\includegraphics[width=0.157\textwidth]{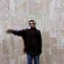}
\caption{\textbf{Weizmann}: \textit{One hand waving}}
\label{fig:weizmann_cat1}
\end{subfigure}
\caption{\textbf{Conditional video generation} on MUG and Weizmann. For both datasets, each line is generated with random $z_m$. We observe that same category (\textit{smile} and \textit{one hand waving}) is performed in a different manner, which indicates that our method is able to produce \textit{intra-class} generation. See SM for more samples.}
\label{fig:cond_results}
\end{figure}

\begin{figure}[htbp]
\begin{subfigure}{0.48\textwidth}
\centering
\includegraphics[width=0.157\textwidth]{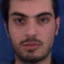}
\includegraphics[width=0.157\textwidth]{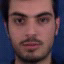}
\includegraphics[width=0.157\textwidth]{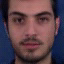}
\includegraphics[width=0.157\textwidth]{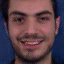}
\includegraphics[width=0.157\textwidth]{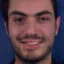}
\includegraphics[width=0.157\textwidth]{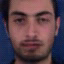}
\includegraphics[width=0.157\textwidth]{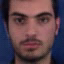}
\includegraphics[width=0.157\textwidth]{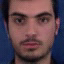}
\includegraphics[width=0.157\textwidth]{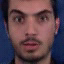}
\includegraphics[width=0.157\textwidth]{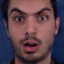}
\includegraphics[width=0.157\textwidth]{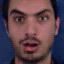}
\includegraphics[width=0.157\textwidth]{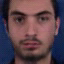}
\caption{\textbf{G$^{3}$AN}}
\label{fig:mug_us}
\end{subfigure}
\begin{subfigure}{0.48\textwidth}
\centering
\includegraphics[width=0.157\textwidth]{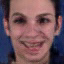}
\includegraphics[width=0.157\textwidth]{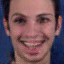}
\includegraphics[width=0.157\textwidth]{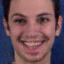}
\includegraphics[width=0.157\textwidth]{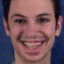}
\includegraphics[width=0.157\textwidth]{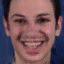}
\includegraphics[width=0.157\textwidth]{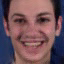}
\includegraphics[width=0.157\textwidth]{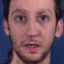}
\includegraphics[width=0.157\textwidth]{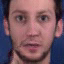}
\includegraphics[width=0.157\textwidth]{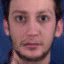}
\includegraphics[width=0.157\textwidth]{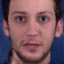}
\includegraphics[width=0.157\textwidth]{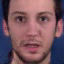}
\includegraphics[width=0.157\textwidth]{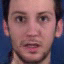}
\caption{\textbf{MoCoGAN}}
\label{fig:mug_mocogan}
\end{subfigure}
\caption{\textbf{Comparison between G$^{3}$AN and MoCoGAN.} Given fixed $z_a$ and $z_m$, as well as two condition-labels \textit{smile} and \textit{surprise}, G$^{3}$AN and MoCoGAN generate correct facial expressions. However, while G$^{3}$AN preserves the appearance between rows, MoCoGAN alters the subject's appearance.}
\label{fig:mug_us_mocogan}
\end{figure}

Further, we leverage on labels of the MUG and Weizmann datasets, in order to analyze \textbf{conditional video generation}. Towards this, we here concatenate a one-hot category vector and motion noise $z_m$, feeding it into $G_T$. We note that the inputs of $G_S$ and $G_V$ remain the same as in the setting of unconditional generation. 
Related results show that when varying motion-categories, while having a fixed appearance, G$^{3}$AN correctly generates an identical facial appearance, with appropriate category-based motion (facial expressions and body actions), see Figure \ref{fig:cond_results}. Further, we note that appearance is very well preserved in different videos and is not affected by category-alterations. In addition, in the same conditional setting, we note that when varying the noise $z_m$, G$^{3}$AN is able to generate \textit{the same category-motion in different ways}. This indicates that $z_m$ enables an intra-class diversity.

In videos generated by MoCoGAN, we observe a correctly generated motion according to given categories, however we note that the category also affects the appearance.
In other words, MoCoGAN lacks a complete disentanglement of appearance and motion in the latent representation, see Figure \ref{fig:mug_us_mocogan}. 
This might be due to a simple motion and content decomposition in the latent space, which after a set of convolutions can be totally ignored in deeper layers. 
It is notable that G$^{3}$AN effectively prevents such cases, ensured by our decomposition that occurs in both, latent and feature spaces.

\begin{figure}[t]
\begin{subfigure}{0.48\textwidth}
\centering
\includegraphics[width=0.157\textwidth]{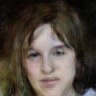}
\includegraphics[width=0.157\textwidth]{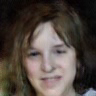}
\includegraphics[width=0.157\textwidth]{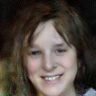}
\includegraphics[width=0.157\textwidth]{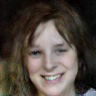}
\includegraphics[width=0.157\textwidth]{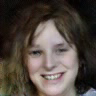}
\includegraphics[width=0.157\textwidth]{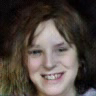}
\includegraphics[width=0.157\textwidth]{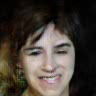}
\includegraphics[width=0.157\textwidth]{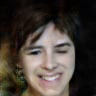}
\includegraphics[width=0.157\textwidth]{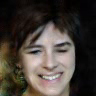}
\includegraphics[width=0.157\textwidth]{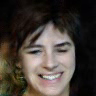}
\includegraphics[width=0.157\textwidth]{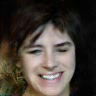}
\includegraphics[width=0.157\textwidth]{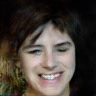}
\caption{Manipulation of \textit{third} dimension on UvA-NEMO}
\label{subfig:uva_dimesnion3}
\end{subfigure}
\begin{subfigure}{0.48\textwidth}
\centering
\includegraphics[width=0.157\textwidth]{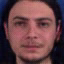}
\includegraphics[width=0.157\textwidth]{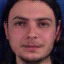}
\includegraphics[width=0.157\textwidth]{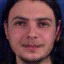}
\includegraphics[width=0.157\textwidth]{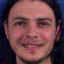}
\includegraphics[width=0.157\textwidth]{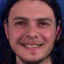}
\includegraphics[width=0.157\textwidth]{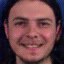}
\includegraphics[width=0.157\textwidth]{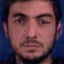}
\includegraphics[width=0.157\textwidth]{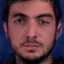}
\includegraphics[width=0.157\textwidth]{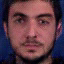}
\includegraphics[width=0.157\textwidth]{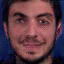}
\includegraphics[width=0.157\textwidth]{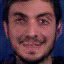}
\includegraphics[width=0.157\textwidth]{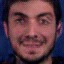}
\caption{Manipulation of \textit{third} dimension on MUG}
\label{subfig:mug_dimension2}
\end{subfigure}
\begin{subfigure}{0.48\textwidth}
\centering
\includegraphics[width=0.157\textwidth]{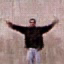}
\includegraphics[width=0.157\textwidth]{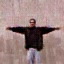}
\includegraphics[width=0.157\textwidth]{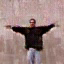}
\includegraphics[width=0.157\textwidth]{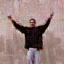}
\includegraphics[width=0.157\textwidth]{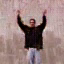}
\includegraphics[width=0.157\textwidth]{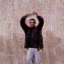}
\includegraphics[width=0.157\textwidth]{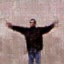}
\includegraphics[width=0.157\textwidth]{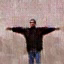}
\includegraphics[width=0.157\textwidth]{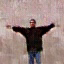}
\includegraphics[width=0.157\textwidth]{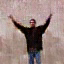}
\includegraphics[width=0.157\textwidth]{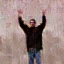}
\includegraphics[width=0.157\textwidth]{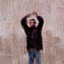}
\caption{Manipulation of \textit{second} dimension on Weizmann}
\label{subfig:uva_dimension4}
\end{subfigure}
\caption{\textbf{Latent appearance representation manipulation.} 
For each dataset, each row shares the same motion representation, whereas from top to bottom values in one dimension of appearance representation are increased. See SM for more samples.}
\label{fig:manipulate_appearance}
\end{figure}

\begin{figure}[htbp]
\begin{subfigure}{0.48\textwidth}
\centering
\includegraphics[width=0.157\textwidth]{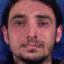}
\includegraphics[width=0.157\textwidth]{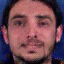}
\includegraphics[width=0.157\textwidth]{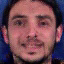}
\includegraphics[width=0.157\textwidth]{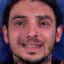}
\includegraphics[width=0.157\textwidth]{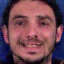}
\includegraphics[width=0.157\textwidth]{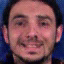}
\includegraphics[width=0.157\textwidth]{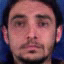}
\includegraphics[width=0.157\textwidth]{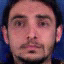}
\includegraphics[width=0.157\textwidth]{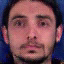}
\includegraphics[width=0.157\textwidth]{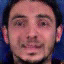}
\includegraphics[width=0.157\textwidth]{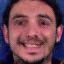}
\includegraphics[width=0.157\textwidth]{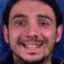}
\caption{Manipulation of \textit{sixth} dimension of MUG}
\label{subfig:motion_mug}
\end{subfigure}
\begin{subfigure}{0.48\textwidth}
\centering
\includegraphics[width=0.157\textwidth]{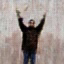}
\includegraphics[width=0.157\textwidth]{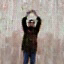}
\includegraphics[width=0.157\textwidth]{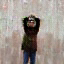}
\includegraphics[width=0.157\textwidth]{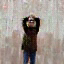}
\includegraphics[width=0.157\textwidth]{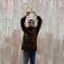}
\includegraphics[width=0.157\textwidth]{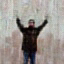}
\includegraphics[width=0.157\textwidth]{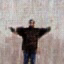}
\includegraphics[width=0.157\textwidth]{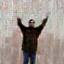}
\includegraphics[width=0.157\textwidth]{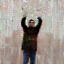}
\includegraphics[width=0.157\textwidth]{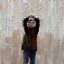}
\includegraphics[width=0.157\textwidth]{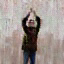}
\includegraphics[width=0.157\textwidth]{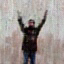}
\caption{Manipulation of \textit{second} dimension on Weizmann}
\label{subfig:weizmann_dim2}
\end{subfigure}
\caption{\textbf{Latent motion representation manipulation.} For each dataset, each row shares the same appearance representation, whereas from top to bottom values in one dimension of the motion representation are increased. See SM for more results.}
\label{fig:manipulate_motion}
\end{figure}

\textbf{Latent representation manipulation.} While there is currently no general method for quantifying the degree of learnt disentanglement \cite{higgins2017beta}, we proceed to illustrate the ability of our model to learn latent representations by manipulating each dimension in the appearance representation. 
We show that by changing \textit{values} of different dimensions in the \textit{appearance representation}, we are able to cause a modification of specific appearance factors, see Figure \ref{fig:manipulate_appearance}. Interestingly such factors can be related to semantics, \eg facial view point in Figure \ref{subfig:uva_dimesnion3}, mustache in Figure \ref{subfig:mug_dimension2}, and color of pants in Figure \ref{subfig:uva_dimension4}.
We note that motion is not affected by altering the appearance representation. 
Similarly, when altering \textit{values} of different dimensions in the \textit{motion representation}, we observe that factors such as starting position, motion intensity and moving trajectory are being affected, see Figure \ref{fig:manipulate_motion}.
Such observations show that our method learns to interpolate between different data points in motion- and appearance-latent spaces, respectively. 


\begin{figure}[htbp]
\begin{subfigure}{0.48\textwidth}
\centering
\includegraphics[width=0.157\textwidth]{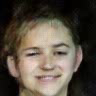}
\includegraphics[width=0.157\textwidth]{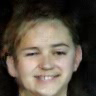}
\includegraphics[width=0.157\textwidth]{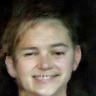}
\includegraphics[width=0.157\textwidth]{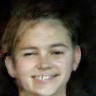}
\includegraphics[width=0.157\textwidth]{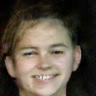}
\includegraphics[width=0.157\textwidth]{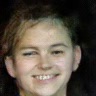}
\includegraphics[width=0.157\textwidth]{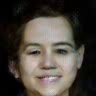}
\includegraphics[width=0.157\textwidth]{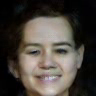}
\includegraphics[width=0.157\textwidth]{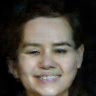}
\includegraphics[width=0.157\textwidth]{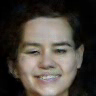}
\includegraphics[width=0.157\textwidth]{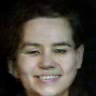}
\includegraphics[width=0.157\textwidth]{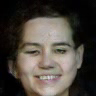}
\caption{$z_{a_0}$, $z_{m_0}$ (top) and $z_{a_2}$, $z_{m_0}$ (bottom)}
\label{subfig:vector_m1}
\end{subfigure}
\begin{subfigure}{0.48\textwidth}
\centering
\includegraphics[width=0.157\textwidth]{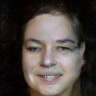}
\includegraphics[width=0.157\textwidth]{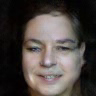}
\includegraphics[width=0.157\textwidth]{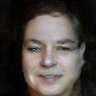}
\includegraphics[width=0.157\textwidth]{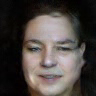}
\includegraphics[width=0.157\textwidth]{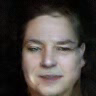}
\includegraphics[width=0.157\textwidth]{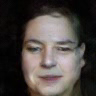}
\includegraphics[width=0.157\textwidth]{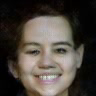}
\includegraphics[width=0.157\textwidth]{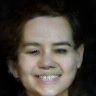}
\includegraphics[width=0.157\textwidth]{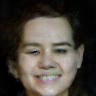}
\includegraphics[width=0.157\textwidth]{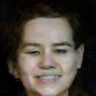}
\includegraphics[width=0.157\textwidth]{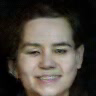}
\includegraphics[width=0.157\textwidth]{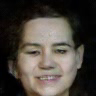}
\caption{$z_{a_1}$, $z_{m_1}$ (top) and $z_{a_2}$, $z_{m_1}$ (bottom)}
\label{subfig:vector_m2}
\end{subfigure}
\caption{\textbf{Addition of appearance representations.} We add the appearance vectors of two samples (top rows of (a) and (b)), and obtain the sum-appearance in each bottom row. We inject motion pertained to each top appearance of (a) and (b) and are able to show same motion within lines of (a) and (b).}
\label{fig:vector_appearance}
\end{figure}

\textbf{Addition of appearance representations.} We here \textit{add} appearance vectors, aiming to analyze the resulting latent representations. Towards this, we generate two videos $V_a$ and $V_b$ by randomly sampling two sets of noises, ($z_{a_0}$, $z_{m_0}$) and ($z_{a_1}$, $z_{m_1}$). Next, we add $z_{a_0}$ and $z_{a_1}$, obtaining a new appearance $z_{a_2}$. When combining ($z_{a_2}$, $z_{m_0}$) and ($z_{a_2}$, $z_{m_1}$), we observe in the two new resulting videos a \textit{summary appearance} pertaining to $z_{a_0}$ and $z_{a_1}$, with identical motion as $z_{m_0}$ and $z_{m_1}$, see Figure \ref{fig:vector_appearance}.

\subsubsection{Ablation Study}
\begin{figure}[htb]
\begin{subfigure}{.235\textwidth}
\centering
\includegraphics[width=0.32\textwidth]{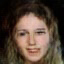}
\includegraphics[width=0.32\textwidth]{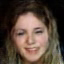}
\includegraphics[width=0.32\textwidth]{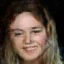}
\includegraphics[width=0.32\textwidth]{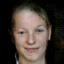}
\includegraphics[width=0.32\textwidth]{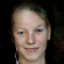}
\includegraphics[width=0.32\textwidth]{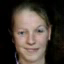}
\includegraphics[width=0.32\textwidth]{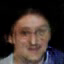}
\includegraphics[width=0.32\textwidth]{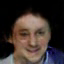}
\includegraphics[width=0.32\textwidth]{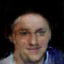}
\caption{$z_a$, $z_{m_0}$}
\label{fig:sub-uncond-a}
\end{subfigure}
\begin{subfigure}{.235\textwidth}
\centering
\includegraphics[width=0.32\textwidth]{demo1/sub3/motion1/frames_0001.png}
\includegraphics[width=0.32\textwidth]{demo1/sub3/motion1/frames_0007.png}
\includegraphics[width=0.32\textwidth]{demo1/sub3/motion1/frames_0016.png}
\includegraphics[width=0.32\textwidth]{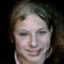}
\includegraphics[width=0.32\textwidth]{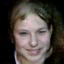}
\includegraphics[width=0.32\textwidth]{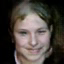}
\includegraphics[width=0.32\textwidth]{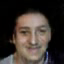}
\includegraphics[width=0.32\textwidth]{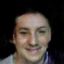}
\includegraphics[width=0.32\textwidth]{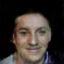}
\caption{$z_a$, $z_{m_1}$}
\label{fig:sub-uncond-b}
\end{subfigure}
\caption{\textbf{Ablation study.} Generated videos obtained by removing $G_T$ (\textit{top row}), removing $G_S$ (\textit{middle}), and both (\textit{bottom row}).}
\label{fig:ablation-1}
\end{figure}

We here seek to study the effectiveness of proposed G$^{3}$AN architecture, as well as the effectiveness related to each component in the proposed Generator. Towards this, we firstly generate videos by removing $G_S$ and $G_T$, respectively, in order to verify their ability of controlling motion and appearance. We observe that when removing $G_T$, the model is able generate different subjects, however for each person the facial movement is constant, see Figure \ref{fig:ablation-1} (\textit{top}). Similarly, when $G_S$ is removed, changing motion will affect subject's identity, whereas the appearance vector loses its efficacy, see Figure \ref{fig:ablation-1} (\textit{middle}). When removing both, $G_T$ and $G_S$, appearance and motion are entangled and they affect each other, see Figure \ref{fig:ablation-1} (\textit{bottom}). This demonstrates the effective disentanglement brought to the fore by the streams $G_S$ and $G_T$.

We proceed to demonstrate the contribution of $G_S$, $G_T$ and F-SA in the Generator \wrt video quality. In this context, we remove each component individually and report results on the four datasets in Table \ref{tab:ablation study}. The results show that after removing all three components, video quality is the poorest, which proves that all of them contribute to the final results. Individually, $G_S$ plays the most pertinent role, as removing it, decreases FID most profoundly for all datasets. This indicates that generating appearance features separately can be instrumental for good quality videos. Moreover, our results confirm the necessity of F-SA in our approach.


\begin{table}[htb]
\setlength{\tabcolsep}{2.1pt}
\centering
\begin{tabular}{cccccc}
\hline
\textbf{Architecture} & \textbf{MUG} & \textbf{UvA} & \textbf{Weizmann} &
\multicolumn{2}{c}{\textbf{UCF101}} \\
& FID $\downarrow$ & FID $\downarrow$ & FID $\downarrow$ & FID $\downarrow$ & IS $\uparrow$ \\
\hline
w/o $G_S$,$G_T$,F-SA & 117.10 & 164.04 & 252.97 & 127.09 & 2.78\\
w/o $G_S$,$G_T$ & 113.44 & 159.54 & 176.73 & 120.17 & 3.16 \\
w/o $G_S$ & 109.87 & 129.84  & 141.06  & 117.19 & 3.05 \\
w/o F-SA & 85.11 & 128.14 & 97.54 & 98.37 & 3.44 \\
w/o $G_T$ & 82.07 & 121.87 & 94.64 & 96.47 & 3.16 \\
G$^{3}$AN & 67.12 & 119.22 & 86.01 & 91.21 & 3.62\\
\hline
\end{tabular}
\caption{\textbf{Contribution of main components in $G$.}}\label{tab:ablation study}
\end{table}


\textbf{Transposed Convolutions.} Then, we compare the proposed factorized transposed spatio-temporal (1+2)D convolution, standard transposed 3D convolution, and transposed (2+1)D convolution, when used in $G_V$ \wrt video quality. We carefully set the number of kernels, allowing for the three networks to have nearly same training parameters. We report the results of the quantitative evaluation in Table \ref{tab:convolution_type}. Both convolution types, (1+2)D and (2+1)D outperform standard 3D kernels \wrt generated video quality. (1+2)D is slightly better than (2+1)D, and the reason might be that the (1+2)D kernel uses more $1\times 1$ kernels to refine temporal information, which we believe to be important in video generation tasks.        

\begin{table}[htb]
\setlength{\tabcolsep}{3.2pt}
\centering
\begin{tabular}
{cccccc}
\hline
&{\textbf{MUG}} &
{\textbf{UvA}} &
{\textbf{Weizmann}} &
\multicolumn{2}{c}{\textbf{UCF101}} \\
& FID $\downarrow$ & FID $\downarrow$ & FID $\downarrow$ & FID $\downarrow$ & IS $\uparrow$ \\
\hline
3D     & 93.51 & 149.98 & 154.21 & 117.61 & 2.88 \\
(2+1)D & 73.08 & 141.35 & 95.01 & 98.70 & \textbf{3.36} \\
(1+2)D & \textbf{69.42} & \textbf{140.42} &  \textbf{87.04} & \textbf{96.79}  & 3.07 \\
\hline
\end{tabular}
\caption{\textbf{Comparison of various convolution types in $G$}.}
\label{tab:convolution_type}
\end{table}{}



\textbf{Where to insert self-attention?} Finally, we proceed to explore at which level of the Generator, F-SA is the most effective. We summarize performance rates in Table~\ref{tab:attention_layers}. Inserting F-SA after the G$^3_3$ module provides the best results, which indicates that \textit{middle level feature maps} contribute predominantly to video quality. As shown in GAN Dissection~\cite{bau2019gandissect}, \textit{mid-level features} represent semantic information, \eg, object parts while \textit{high-level features} represent local pixel patterns, \eg, edges, light and colors and \textit{low-level features} do not contain clear semantic information, which could be the reason, why F-SA achieves the best result in G$^3_3$ module.


\begin{table}[htb]
\centering
\begin{tabular}
{cccccc}
\hline
&{\textbf{MUG}} &
{\textbf{UvA}} &
{\textbf{Weizmann}} &
\multicolumn{2}{c}{\textbf{UCF101}} \\
& FID $\downarrow$ & FID $\downarrow$ & FID $\downarrow$ & FID $\downarrow$ & IS $\uparrow$ \\
\hline
G$^{3}_0$ & 83.01 & 188.60 & 96.38 & 100.37 & 3.09 \\
G$^{3}_1$ & 72.54 & 178.64 & 99.66 & 126.12 & 2.74 \\
G$^{3}_2$ & 69.02 & 160.12 & 97.53 & 112.36 & 3.03 \\
G$^{3}_3$ & \textbf{67.12} & \textbf{119.22} & \textbf{86.01} & \textbf{91.21} & \textbf{3.62} \\
\hline
\end{tabular}
\caption{\textbf{Comparison of inserting F-SA at different hierarchical levels of G$^3$AN.}}
\label{tab:attention_layers}
\end{table}

\section{Conclusions}\label{conclusion}
We have presented the novel video generative architecture G$^3$AN, 
which leverages among others on (i) a three-stream Generator that models appearance and motion in disentangled manner, as well as (ii) a novel spatio-temporal fusion method. We have performed an extensive evaluation of our approach on four datasets, outperforming quantitatively and qualitatively the state-of-the-art video generation methods VGAN, TGAN and MoCoGAN. Further, we have shown the ability of G$^3$AN to disentangle appearance and motion, and hence manipulate them individually. 

Future work involves the design of GAN models for video generation of high resolution videos. 
\paragraph{Acknowledgment.} This work is supported by the French Government (National Research Agency, ANR), under Grant ANR-17-CE39-0002.




{\small
\bibliographystyle{ieee_fullname}
\balance
\bibliography{main}
}

\end{document}